# Beyond Local Search: Tracking Objects Everywhere with Instance-Specific Proposals

Gao Zhu[1], Fatih Porikli[1,2,3], and Hongdong Li[1,3]
Australian National University[1] and NICTA[2]
ARC Centre of Excellence for Robotic Vision[3]
{gao.zhu,fatih.porikli,hongdong.li}@anu.edu.au *

## Abstract

*Most tracking-by-detection methods employ a local search window around the predicted object location in the current frame assuming the previous location is accurate, the trajectory is smooth, and the computational capacity permits a search radius that can accommodate the maximum speed yet small enough to reduce mismatches. These, however, may not be valid always, in particular for fast and irregularly moving objects. Here, we present an object tracker that is not limited to a local search window and has ability to probe efficiently the entire frame. Our method generates a small number of "high-quality" proposals by a novel instance-specific objectness measure and evaluates them against the object model that can be adopted from an existing tracking-by-detection approach as a core tracker. During the tracking process, we update the object model concentrating on hard false-positives supplied by the proposals, which help suppressing distractors caused by difficult background clutters, and learn how to re-rank proposals according to the object model. Since we reduce significantly the number of hypotheses the core tracker evaluates, we can use richer object descriptors and stronger detector. Our method outperforms most recent state-of-the-art trackers on popular tracking benchmarks, and provides improved robustness for fast moving objects as well as for ultra low-frame-rate videos.*

## 1. Introduction

Model-free object tracking, which aims to track arbitrary objects based on a single bounding-box annotation, has gained significant attention recently with numerous approaches [22, 16, 18] proposed and several large benchmark datasets [39, 32, 26, 38] released. Significant amount of effort has been devoted to either designing a better object representation, including subspace [30], sparsity [28, 43], and deep learned features [34, 41], or building complex classifiers [16, 4, 45, 40, 29] for better discrimination of the object from its background patches. Most of these methods, however, require a search window centered at the previous object location to select candidate patches, partly due to computational complexity. This is sometimes referred as the motion model [35], and it is implicitly assumed that the object is correctly tracked in the previous frames and the object motion is not large. Even though this simplification works in some situations, it also introduces serious difficulties especially when the object undergoes deformations and occlusions (which may cause drift), or when the object and camera motion puts the object beyond the search window radius.

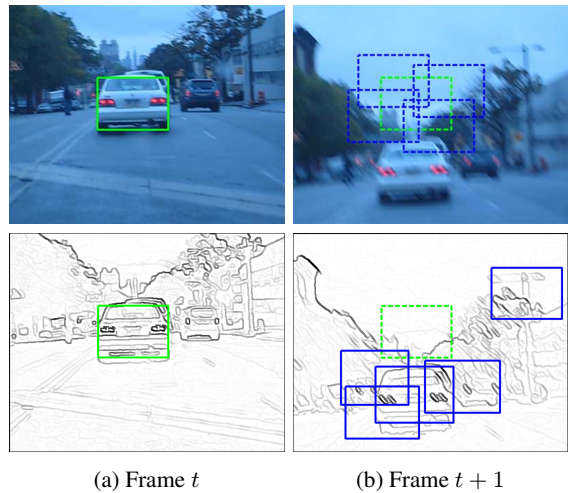

(a) Frame $t$  (b) Frame $t+1$

Figure 1: **Top row:** Most existing tracking-by-detection methods examine hypothesis locations within a local and heuristically defined search window around the last detected location. **Bottom row:** Our tracker seeks high-quality hypotheses over the entire image using instance-specific edge-box locations.

*This work was supported under the Australian Research Councils Discovery Projects funding scheme (project DP150104645, DP120103896), Linkage Projects funding scheme (LP100100588), ARC Centre of Excellence on Robotic Vision (CE140100016) and NICTA (Data61).

One important reason that the existing trackers avoid employing a wider search radius is the potential distractions from the background [13, 29]. It is not a trivial task to update a discriminative classifier when the negative sample space grows greatly with the samples coming from the extended search radius. In [18], extended set of training data is obtained by implicitly including all shifted versions of the given samples within the circulant matrices. However, it is impractical to apply the same trick for the negative samples, especially for the ones far away from the object.

To overcome this, in this work we introduce a proposal generation procedure for handling the problem of sample selection, both for the object detection and the model update stages. Generally, the motion model limits the search radius and the applied sampling schemes disregard the contents presented on them. Instead of working within a limited search radius, we generate a small yet high-quality set of proposals efficiently in entire frame by using simple bottom-up, edge-based features [46] as shown in Figure 1. Intuitively, edge information provides valuable guidance for object tracking since objects may often be identified by their silhouettes. In addition, concentrating on image regions where edge information is eminent allows efficient selection of more object-like proposals.

Our method can incorporate any existing object model including simpler template matching models, e.g. normalized cross correlation (NCC) and sophisticated classifiers, e.g., structured support vector machines (SSVM). Using the object model, we adapt the edge-based features used in proposal generation. In an online fashion, we learn how to re-rank the proposal by a linear support vector machine, trained on the current proposals, with a crafted feature vector. Our proposal scheme, thus, generates windows that suggest certain similarity to the tracked object. This allows taking advantage of objectness to regulate the proposal selection in a temporally coherent manner instead of treating objectness as yet another cue by (linearly) combining the original tracking response with some objectness score. Since we adapt the generic edge-based objectness measure to the specific object, this selection is superior to replacing the search window with simple objectness responses.

Furthermore, for the chosen object model, we explore the best combination of global proposals provided by instance specific edge-based features and local candidates sampled around the previous location for model update (e.g., for negative support vectors in case of SSVM). We also adapt the size and scale to obtain the best proposals.

The benefits of our proposal generation is threefold:

- Our method can execute global search over entire image. Thus, it can track objects without making any assumption on object motion.
- The high-quality proposals increase the tracking accuracy since they allow including better hard negatives into training set, hence reduces drift.
- It adapts the specific object, thus provides better object model update (than generic proposals).

We validate the above arguments with two object models (from NCC tracker and Struck) and show that the incorporation of instance-specific proposals has potential to improve most detection-by-tracking approaches.

Our method is conceptually simple, easy to implement, and most importantly, provides the best results (at the time of submission) in comparison to all state-of-the-art trackers. Our method ranks as the top tracker on VOT2014 [26] benchmark as well as on OTB [39] and TB50 [38] datasets in comparison to the latest state-of-the-art including MEEM [40], KCF [18], Struck [16], and over twenty other methods.

## 2. Related Work

Providing an inclusive overview of the object tracking literature is outside the scope and capacity of this paper. We refer readers to the excellent surveys on object tracking. Here, we only compare with some relevant algorithms. We briefly examine different search schemes and then summarize recent object proposal methods.

**Search Schemes in Tracking**

There is a wide-spectrum of styles to select which windows will be tested in a current frame to locate the target object and also update its model.

**Single Window Search:** Several trackers use the local window around the former object location to find the object in the current frame. Examples include the tracking on Lie groups [33], which applies iteratively a feature-motion regressor to estimate object window in the next frame, and the mean-shift tracker [11], which uses gradient-based local optimization to determine the mode of the underlying similarity distribution.

**Particle-based Search:** In recent years, tracking algorithms [30, 44, 21] based on particle filtering has been extensively studied. Particle filters apply importance sampling on the previous particle states (e.g. candidate locations) within mostly a mixed number of candidates. On the negative side, the random sampling is blind to the underlying texture, edgeness, and other spatial information.

**Searching for the Hard Negatives:** It is worthwhile to mention that tracking-by-detection, which allows an online trained classifier [3, 31] as an object model to distinguish the object from its surrounding background, has recently become particularly popular. Rather than explicitly coupling to the accurate estimation of object position, [4] limits its focus on increasing the robustness to poorly labeled samples. [16] proposes directly predicting the change in object location between frames by an online structured output SVM. Even though it produces comparably accurate tracking, it uniformly samples the state space to generate positive and negative support vectors. Such a brute force approach on a larger search window is computationally intractable.

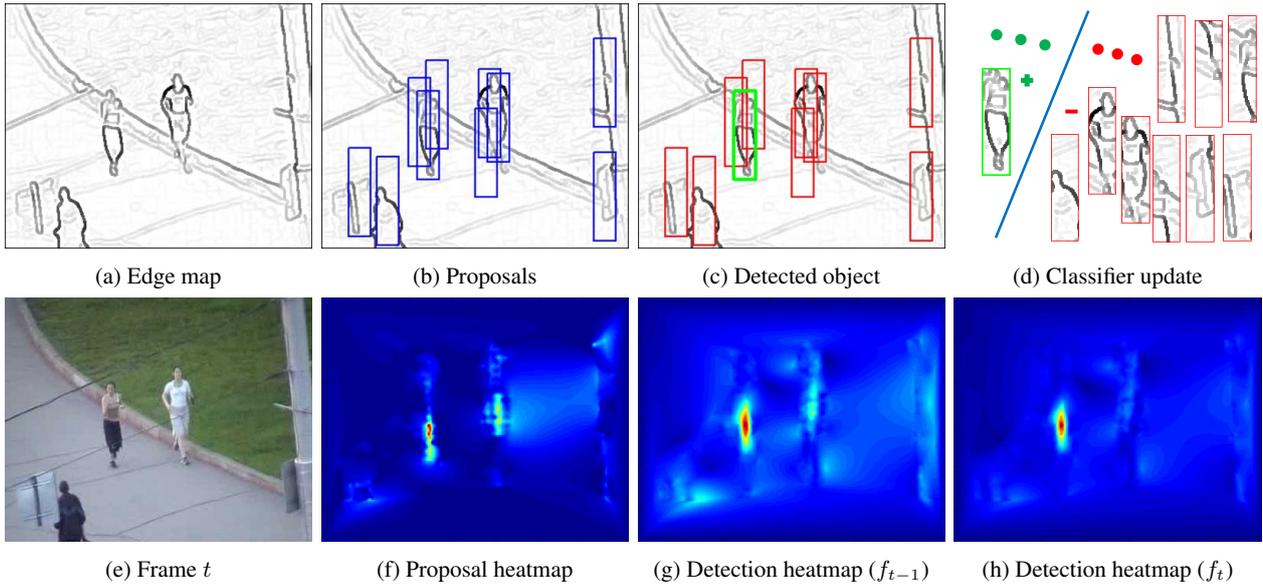

Figure 2: Framework of the proposed method. First column: (a) Edge map extracted from the current frame (e); Second column: (b) Object proposals in blue bounding boxes (Section 3.3) and (f) corresponding heatmap of instance specific proposals; Third column: (c) Detection results on proposals (green is detected as object) and (g) detection heatmap (by the proposed EBT classifier); Fourth column: (d) EBT is updated using the proposals and (h) detection heatmap with updated EBT. Notice that spurious hypotheses (bright regions in (g)) are suppressed significantly by treating them as negative samples.

**Objectness in Object Detection**

As shown in [19, 46], use of proposal has significantly improved the object detection benchmark along with the convolutional neural nets. Since, a subset of high-quality candidates are used for detection, object proposal methods improve not only the speed but also the accuracy by reducing false positives. The top performing detection methods [15, 36] for PASCAL VOC [14] use detection proposals.

**Edge Box:** [46] proposes object candidates based on the observation that the number of contours wholly enclosed by a bounding box is an indicator of the likelihood of the box containing an object. Edge Box is designed as a fast algorithm to balance between speed and proposal recall. Its 1-D feature generates remarkably accurate results.

**BING:** [10] made a similar observation that generic objects with well-defined closed boundary can be discriminated by looking at the norm of gradients. They further designed a feature called binarized normed gradients (BING), which can be used for efficient objectness estimation and requires only a few atomic operations.

**Objectness as Supportive Cue for Tracking**

A straightforward strategy, i.e., linear combination of the original tracking confidence and an adaptive objectness score based on BING [10] is employed in [25]. In [20], a detection proposal scheme is applied as a post-processing step, mainly to improve the tracker's adaptability to scale and aspect ratio changes. These methods are substantially different from our work, where we adapt objectness to specific object using a separate classifier and generate high-quality proposal to regulate the tracking process.

## 3. Global Tracking with Proposals

### 3.1. Pipeline

A typical tracking-by-detection framework is composed mainly of motion model, observation model and model updater [39, 32, 35]. Motion model generates a set of candidates which might contain the target in the current frame based on the estimation from the previous frame. Observation model judges whether a candidate is the target based on the features extracted from it. Model updater online updates the observation model to adapt the change of the object appearance.

Suppose the object location is initialized manually at the first frame $t = 1$ and $B_t$ is its bounding box at frame $t$. Then, given an observation model, i.e., a classification function $f_{t-1}$ trained on the previous frames, the current loca-

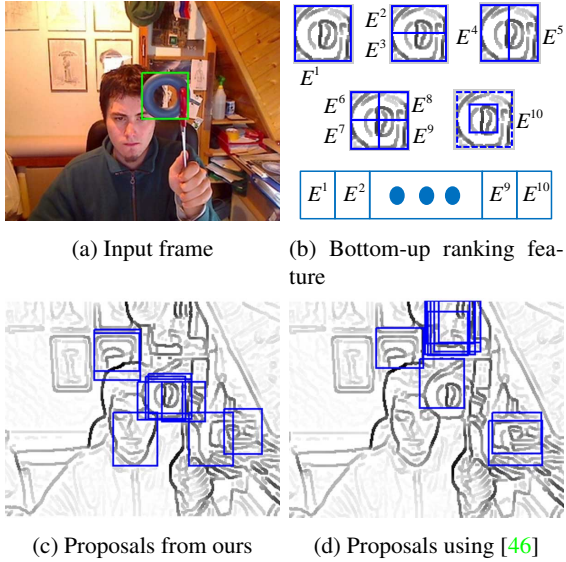

Figure 3: Instance specific proposals. (a) Input frame (ground truth is the green bounding box); (b) 10-dimensional feature vector for ranking of the bounding boxes; (c) Top proposals using the proposed method; (d) Top proposals from [46]. As shown, the instance specific proposals are far more precise.

tion of the object is estimated through:

$$B_t^\star = \arg\max_{B_t \in \mathcal{B}_t} f_{t-1}(B_t), \qquad (1)$$

where $\mathcal{B}_t$ is a set of samples generated by the motion model at the current frame. To select samples, traditional trackers use heuristic search windows around the previously estimated object location for computational and accuracy reasons. For example, a search radius of 30 pixels is used in [16].

Each sample is labeled by a classifier that models the object. The update routine will then revises its model $f_{t-1} \to f_t$ with the new location of the object to adapt possible appearance changes. It is not trivial to design a robust updating scheme [27, 35]. As there is only one reliable example, the tracker must maintain a trade-off between adapting to new but possibly noisy examples collected during tracking and preventing the tracker from drifting to the background.

### 3.2. Our Method

The method proposed in this paper uses a similar framework as introduced in Section 3.1, yet we made two critical changes to the motion model. The first change is that we recognize not all candidate bounding boxes $B_t \in \mathcal{B}_t$ should be treated equally (as the traditional trackers often do) since those boxes possess different *object-like* appearance, i.e. *objectness* [2, 9] characteristics, which should be taken into account. Secondly, we do not constrain the search radius to a small window that causes throwing so much available image information away.

To execute our changes, we take advantage of the sparse, simple, yet critical edge information. The current frame $I_t$ is processed into an edge map as shown in Figure 2a. Then, we employ an instance specific proposal method (explained in Section 3.3) build on top of the object proposal algorithm [46] to produce a number of candidate bounding boxes (Figure 2b and 3c) denoted as $\mathcal{B}_t^E$. Notice that, we impose a smooth size change constraint to the bounding boxes between consecutive frames.

Suppose the bounding box set generated by sampling only around the previous object location as $\mathcal{B}_t^R$ (as in traditional methods). Now we have two different sets of candidates, i.e., $\mathcal{B}_t^E$ and $\mathcal{B}_t^R$. The first one possesses object regularity while the second one is with no discriminative information. As shown in the experimental section 5.2, the choice of using only the proposals $\mathcal{B}_t^E$ generates the best results, better than combining them together. This confirms our argument that object proposals not only reduce the candidate sample space but also reduce spurious false positive and improve tracking accuracy. Our tracker will not drift to a textureless region like other trackers due to the *objectness* constraint.

During the update stage, we also have different options for using $\mathcal{B}_t^E$ and $\mathcal{B}_t^R$. As validated in the experimental part 5.2, the combination of using both of them to choose negative support vectors results in the best performance. This can be easily explained: $\mathcal{B}_t^E \setminus B_t^\star$ only represents other good *object-like* regions. By putting them as negative support vectors, we would only increase the discriminative power among *objects-like* candidates. However, the negative sample space contains a lot more other negative samples. Thus, the advantageous option is to augment $\mathcal{B}_t^E \setminus B_t^\star$ with $\mathcal{B}_t^R$ in order to achieve the best discriminative ability.

### 3.3. Instance Specific Proposals

Objectness attempts to generate quickly as few as possible hypotheses yet cover all of the objects present in an image. Take EdgeBox [46] for example - it generates a pool of bounding boxes $\{B_{t,i}\}$ uniformly sampled in a sliding window manner, then ranks and extracts the top $H$ candidates with the highest *objectness* score $E_{t,i}$, represented by:

$$\mathcal{B}_t^{EB} = \{B_{t,i}|E_{t,i}\}_H. \qquad (2)$$

$E_{t,i}$ is basically a weighted and normalized number of contours wholly enclosed by the bounding box $B_{t,i}$. This feature can be calculated very efficiently in real-time. We refer [46] for more details.

Instead of directly applying the computed proposals $\mathcal{B}_t^{EB}$ for tracking, we argue that the object instance level properties should be taken into account. As such, there is a strong object prior in terms of its geometric structure of contours and size in contrast to object detection where the goal is to locate all instances of all object classes in the image. EdgeBox generates proposals that favors bounding boxes with many internal contour segments, thus it is likely to miss the target in a cluttered background as shown in Figure 3d.

To this end, we incorporated an online updated linear SVM [37] classifier $f_{t-1}^R$ to re-rank proposals and determine the top $H$ proposals based on their classification scores:

$$\mathcal{B}_t^E = \{B_{t,i} | f_{t-1}^R(B_{t,i})\}_H, \quad (3)$$

with a 10-dimensional feature vector $\{E_{t,i}^1, \ldots, E_{t,i}^{10}\}$ as shown in Figure 3b. This feature characterizes the spatial structure of edge information. It concatenates Edge-Box scores corresponding to Haar wavelet like partitioning of the bounding box $B_{t,i}$. Notice that, only the bounding boxes whose initial *objectness* scores are above a threshold, i.e., $\mathcal{B}_t^{EB_T} = \{B_{t,i} | E_{t,i} > e_T\}$ (in all experiments $e_T = 0.005$) are accepted into the classifier for re-ranking to save computing time.

The re-ranking classifier is initialized using the top EdgeBox proposal (top 200 in all experiments) and then online updated at every 5 frames with the same number of proposals. The estimated position gives the positive sample and bounding boxes which overlap the estimation less than 0.5 are assigned as negative ones. We use the implementation and parameters as in [40].

### 3.4. Candidate Classification

We use the following decision function to estimate the new location of the object (Figure 2c):

$$B_t^\star = \arg\max_{B_t \in \mathcal{B}_t} f_{t-1}(B_t) + s(B_t, B_{t-1}^\star). \quad (4)$$

$s(B_t, B_{t-1}^\star)$ is a term representing the motion smoothness between the previous object location and the candidate box. This is important in our formulation as we are testing candidates all over the image, though not penalizing it too much. We use a simple function in this paper: $s(B_t, B_{t-1}^\star) = w_s \exp(-\frac{1}{2\sigma^2} \|c(B_t) - c(B_{t-1}^\star)\|^2)$, where $c(B_t)$ is the center of bounding box $B_t$, $w_s = 0.1$ and $\sigma$ is set as the diagonal length of the initialized bounding box.

## 4. Proposed Trackers

Two core object models are integrated in the proposal tracker. The first one (called as EBT to indicate its relation to EdgeBox) follows a popular structured support vector machine (SSVM) framework [16], which shows good performance on several benchmarks [39, 32]. We additionally incorporated a much simpler, normalized cross correlation (NCC) template matching, called as NCC$_{\textbf{EB}}$, to investigate how much additional performance improvement our method is able to provide.

### 4.1. EBT Tracker

Suppose the support vector set maintained by the SSVM as $\mathcal{V}_{t-1}$ and the classification function can be written as a weighted sum of affinities [5, 16]:

$$f_{t-1}^S(B_t) = \sum_{B_{t-1}^i \in \mathcal{V}_{t-1}} w_{t-1}^i k(B_{t-1}^i, B_t), \quad (5)$$

where $w_{t-1}^i$ is a scalar weight associated with the support vector $B_{t-1}^i$. Kernel function $k(B_{t-1}^i, B_t)$ calculates the affinity between two feature vectors extracted from $B_{t-1}^i$ and $B_t$ respectively. The classifier is updated in an online fashion using [6] with a budget [37]. Intersection kernel is used and other parameters are set same as [16].

To take advantage of the small set of proposals, we use histogram features obtained by concatenating 16-bin intensity histograms from a spatial pyramid of 5 levels and RGB channels separately. At each level $L$, the patch is divided into $L \times L$ cells, resulting in a 2640-D feature vector, comparing to the 480-D feature used in [16], while running at a similar speed. The performance gain of using the richer feature is demonstrated in the experimental section 5.2.

### 4.2. NCC$_{\textbf{EB}}$ Tracker

The classification function for the normalized cross correlation can be written as:

$$f_{t-1}^N(B_t) = \rho(B_t, B_{Temp}), \quad (6)$$

where $\rho$ calculates the normalized cross-correlation coefficient [7] between the candidate patch and the object template. This procedure can be accelerated using the fast Fourier transform (FFT) trick. We compared the proposed NCC$_{\textbf{EB}}$ tracker with instance-specific proposals and fixed template with: (1) NCC, an implementation from [26], uses local exhaustive search, and has no update; and (2) IMP-NCC, an improved NCC version from [26], uses local exhaustive search, online update, and Kalman Filter [23] for trajectory smoothness.

## 5. Experiments

In the first part, we compare our method with the state-of-the-art trackers on benchmark datasets for a general performance evaluation. We also test on fast-motion related categories to put it under the spotlight to understand how well our method can handle the challenging scenarios such as fast moving objects, randomly moving objects, and tracking under low-frame-rate. In the second part, we analyze different components of our method.

### 5.1. Full Benchmark Evaluations

Our method is tested on three large datasets: OTB [39], TB50 [38] and VOT2014 [26]. The first two of these datasets are composed of around 50 sequences each. They are annotated with ground truth bounding boxes and various visual attributes. TB50 is an upgraded version of OTB and contains much more challenging sequences. VOT2014 dataset selectively collects 25 sequences from various datasets and allows the tracker to re-initialize once the tracker drifts away from the object.

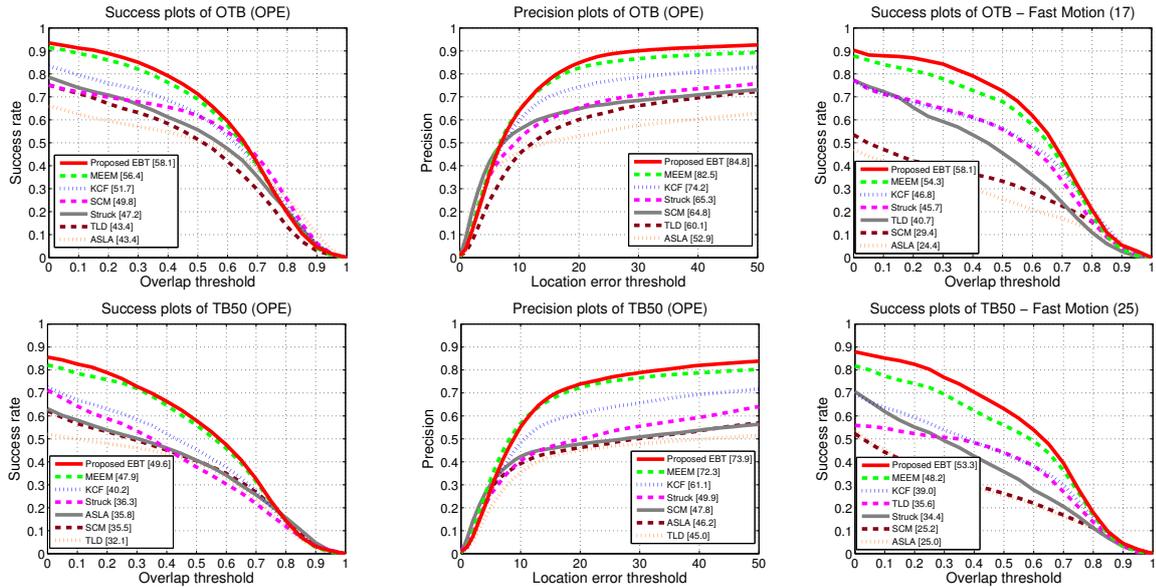

Figure 4: *Success plot* and *precison plot* on two large benchmarks: OTB and TB50. Algorithms are ranked by the area under the curve and the precision score (20 pixels threshold). Our method achieves consistently superior performance.

We compare against the existing algorithms on respective benchmarks and additionally two recent works: KCF [18] and MEEM [40]. Evaluation metrics and code are provided by the respective benchmark. For OTB and TB50, we employ the one-pass evaluation (OPE) and use two metrics: *precision plot* and *success plot*. The former metric calculates the rate of frames whose center location is within a certain threshold distance with the ground truth. The latter one calculates a same ratio but based on bounding box overlap threshold.

**Parameters** For EdgeBox proposals, the sampling step of sliding window is set at $\alpha = 0.85$ since we aim for a high accurate localization. The minimal and maximal areas are $0.5$ and $2$ of the area of the previous estimated bounding box respectively. Non-maximum suppression parameter is fixed at $\beta = 0.8$. The maximum number of proposal is 200 (more discussion in Section 5.2).

### 5.1.1 Benchmark Results

The results are summarized in Table 1, 2 and Figure 4. Our EBT tracker ranks as the best tracker on VOT2014 as shown in Table 1. We use the original VOT protocol. EBT achieves the best overall performance in all datasets[1]. It consistently outperforms the state-of-the-art trackers and improves the base Struck tracker by a large margin. A few examples can be found in Figure 5.

Even the proposed $NCC_{EB}$ tracker using only template matching manages to improve the simple NCC tracker significantly and outperforms several other trackers including the IMPNCC tracker, which has incorporated sophisticated mechanisms in comparison to ours and NCC. This result is not surprising since the incorporation of objectness has proven to be a successful strategy in single image object detection [15, 36, 14]. We believe that our method is a counterpart in the tracking domain as no existing tracking methods successfully adopted such objectness schemes before, to the best of our knowledge.

Table 1: Performance on VOT2014.

|  | Final Rank | Acc. Rank | Rob. Rank |
|---|---|---|---|
| Proposed EBT | **13.03** | 15.81 | 10.24 |
| $PLT_{14}$ [26] | 13.75 | 16.66 | 10.84 |
| $PLT_{13}$ [26] | 14.26 | 18.59 | **9.92** |
| DGT [8] | 14.54 | 15.48 | 13.61 |
| DSST [12] | 15.25 | 13.40 | 17.09 |
| KCF [18] | 15.25 | **12.20** | 18.29 |
| SAMF [24] | 15.47 | 12.79 | 18.15 |
| MEEM [40] | 18.95 | 21.15 | 16.76 |
| Struck [16] | 22.83 | 22.30 | 23.36 |
| Proposed $NCC_{EB}$ | 27.27 | 24.20 | 30.35 |
| MIL [4] | 27.69 | 31.24 | 24.14 |
| FSDT [26] | 27.86 | 25.97 | 29.75 |
| IMPNCC [26] | 27.99 | 26.05 | 29.94 |
| CT [42] | 28.26 | 29.14 | 27.38 |
| FRT [1] | 28.64 | 25.02 | 32.26 |
| NCC [26] | 29.30 | 22.32 | 36.28 |

---
[1]As stated in FAQ of the official VOT website, the rankings would not be identical to the Table 1 in the 2014 paper.

Table 2: Area Under Curve (AUC) of *success plot* and Precision Score (20 pixels threshold) reported on various datasets (AUC/PS) corresponding to the one-pass evaluation (OPE).

|  | Pro. EBT | KCF [18] | MEEM [40] | Struck [16] | SCM [44] | ASLA [21] | TLD [22] | CXT [13] | CSK [17] |
|---|---|---|---|---|---|---|---|---|---|
| OTB | **58.1**/**84.8** | 51.7/74.2 | 56.4/82.5 | 47.2/65.3 | 49.8/64.8 | 43.4/52.9 | 43.4/60.1 | 42.3/57.0 | 39.6/54.1 |
| TB50 | **49.6**/**73.9** | 40.2/61.1 | 47.9/72.3 | 36.3/49.9 | 35.5/47.8 | 35.8/46.2 | 32.1/45.0 | 32.1/43.2 | 31.4/43.0 |

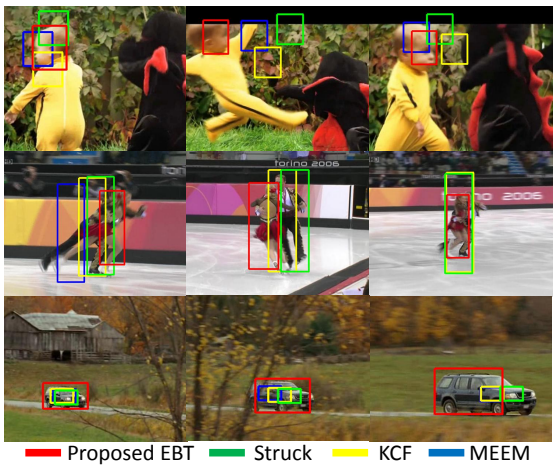

Figure 5: Qualitative comparisons with the state-of-the art trackers on the *DragonBaby*, *Skating2*, and *CarScale* videos. Our method exhibits robustness in challenging scenarios such as fast motion, occlusion, and scale changing.

#### 5.1.2 Tracking Fast Objects

Since our method searches over the entire image, it is suitable for tracking fast moving objects, which could move outside of the search radius of the traditional trackers. As shown in Table 3, our method outperforms other trackers in the fast-motion related categories as well.

We also tested our method on an extra category *Moving Camera* from ALOV300 [32]. This category contains many sequences that depict camera shake, sudden object motion, and abrupt jumps. ALOV300 provides a high number of short sequences with 14 visual attributes. The main source of their data is real-life videos from YouTube.

**Tracking under Ultra-Low-Frame-Rate** We additionally created a dataset, called as VOT2014+ by temporally sampling sequences at every 20 frames on VOT2014, thus, it contains 20× faster moving objects. Our method is tested against with other top-ranked trackers, KCF and MEEM. Even though both MEEM and KCF rapidly failed, our tracker retained **very high** performance scores (see Table.4).

Table 4: Performance on the low-fps dataset.

|  | Pro. EBT | KCF [18] | MEEM [40] |
|---|---|---|---|
| VOT2014 | **46.7**/**65.9** | 38.9/53.7 | 44.5/62.3 |
| VOT2014+ | **43.7**/**58.5** | 28.4/34.1 | 37.5/47.7 |

### 5.2. Further Remarks

**Combination of $\mathcal{B}_t^E$ and $\mathcal{B}_t^R$** As discussed in Section 3.2, we tested different combinations of the hypothesis proposals $\mathcal{B}_t^E$ and candidate bounding boxes $\mathcal{B}_t^R$ sampled around the previous object location within a radius. The results are shown in Table 5. For combinations which use only $\mathcal{B}_t^R$ in the testing stage, we apply an exhaustive sampling within a 30-pixels radius to achieve a comparable result. For the others which use $\mathcal{B}_t^R$, we only generate 80 samples uniformly within a 30-pixels radius. Our main discussion about these results can be found in Section 3.2. We observed the combination of using samples from the hypothesis proposals and local region in update stage and samples only from the proposed locations in the test stage performs the best.

**Number of Proposals** To quantitatively compare the proposed instance specific proposals and the one using Edge Box [46], we analyzed the upper bound performance with respect to varying number of proposals as shown in Figure 6. A variant denoted as EBTeb using EdgeBox proposals instead of ours is also tested and available in Figure 7. Both results show that the proposed re-ranking method outperforms the one directly applies EdgeBox. We also tested the variants using different number of proposals. EBT100 and EBT400 use 100 and 400 respectively, comparing to the proposed EBT that uses 200. Our observations are, using insufficient number of proposal leads to a bad coverage of the false positives as well as the object, while using a large number of proposals attracts spurious candidates.

Table 3: Area Under Curve (AUC) of *success plot* and Precision Score (20 pixels threshold) reported on various fast-motion related categories (AUC/PS). FM: fast motion, MB: motion blur, MC: moving camera. fps: frames per second.

| Attributes | Pro. EBT | KCF [18] | MEEM [40] | Struck [16] | SCM [44] | ASLA [21] | TLD [22] |
|---|---|---|---|---|---|---|---|
| FM (17) (OTB) | **58.1**/**77.8** | 46.8/61.0 | 54.3/71.4 | 45.7/59.6 | 29.4/32.9 | 24.4/24.6 | 40.7/53.2 |
| MB (12) | **58.3**/**77.1** | 50.8/66.0 | 53.0/68.0 | 42.6/54.0 | 29.5/33.3 | 25.1/26.8 | 39.0/49.0 |
| FM (25) (TB50) | **53.3**/**74.5** | 39.0/54.0 | 48.2/68.4 | 34.4/42.5 | 25.2/29.6 | 25.0/29.6 | 35.6/46.5 |
| MB (19) | **54.9**/**78.5** | 40.6/56.4 | 52.8/72.9 | 30.9/35.5 | 21.7/25.1 | 23.3/25.5 | 39.3/49.7 |
| MC (22) (ALOV300) | **60.9**/**68.4** | 56.4/62.9 | 57.2/65.1 | 44.9/44.8 | 35.7/37.9 | 38.6/38.8 | 56.1/67.9 |
| fps | 4.4 | **70.9** | 7.1 | 4.8 | 0.3 | 3.8 | 8.8 |

Table 5: Results for different combinations of $\mathcal{B}_t^E$ and $\mathcal{B}_t^R$.

| TB50 | (Test) $\mathcal{B}_t^R$ | $\mathcal{B}_t^E$ | $\mathcal{B}_t^E + \mathcal{B}_t^R$ |
|---|---|---|---|
| $\mathcal{B}_t^R$ (Update) | 41.1/58.7 | 44.7/64.2 | 42.7/59.4 |
| $\mathcal{B}_t^E$ | 40.1/56.3 | 46.5/68.6 | 43.0/61.8 |
| $\mathcal{B}_t^E + \mathcal{B}_t^R$ | 39.2/56.5 | **49.6**/**73.9** | 43.2/63.6 |

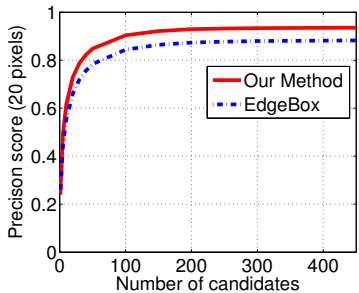

Figure 6: The performance bounds for using EdgeBox proposals and the proposed instance-specific proposal method on TB50. The best candidate in each frame is used for calculating the performance.

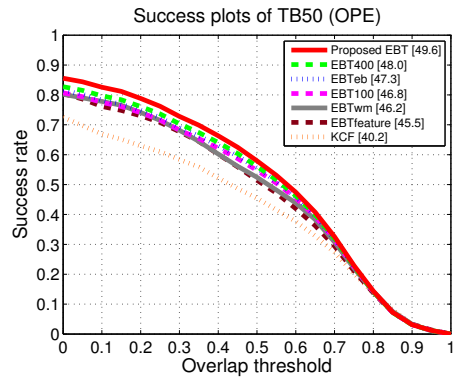

Figure 7: *Success plot* of variants of the proposed method on TB50. Details can be found in Section 5.2.

Table 6: Performance when BING is used instead of Edge Box.

| | Struck [16] | BING-VOC | BING-Adapt |
|---|---|---|---|
| TB50 | **36.3**/**49.9** | 30.8/47.6 | 33.7/48.0 |

**Richer Features and Motion Constraint** EBTfeature denotes the variant using a lower dimensional 480-D feature. This version has lower performance than the one uses 2640-D feature as expected. More details about the feature can be found in Section 4.1. EBTwm denotes the variant without using the smoothness term $s(B_t, B_{t-1}^\star)$ in Function 4. The success rate dropped due to the fact that the motion in the tracking sequences is not completely random.

**Proposals using BING** We evaluated another popular object proposal method, BING [10], for proposals. Two ways of incorporation were tested. The first one (BING-VOC) uses the pretrained model on VOC dataset [14], while the second one (BING-Adapt) relearns the model using the first frame of each sequence. We tested these two variants on TB50. Results are in Table 6. Both performances are worse than the baseline Struck. This is expected. As shown [19, 46], BING results in a relatively low recall of the objects, which is one reason for its mediocre performance.

**Computational Speed** The computational speed of the proposed is comparable to the state-of-the-art trackers even though we can track over the entire image. The proposal part takes less than 100 milliseconds and the overall tracking speed is available in Table 3.

## 6. Conclusion

This paper presented a robust method that can locate objects that are moving randomly and very fast, as well as perform tracking under extremely low-frame rates. To the best of our knowledge, our tracker achieves the **best results on all** common benchmark **datasets** including OTB [39], TB50 [38], VOT2014 [26] and ALOV300 [32].